\documentclass[conference]{IEEEtran}
\IEEEoverridecommandlockouts
\usepackage{cite}
\usepackage{amsmath,amssymb,amsfonts}
\usepackage{algorithmic}
\usepackage{graphicx}
\usepackage{textcomp}
\usepackage{xcolor}
\usepackage{times}
\usepackage{soul}
\usepackage{url}
\usepackage[hidelinks]{hyperref}
\usepackage[utf8]{inputenc}
\usepackage[small]{caption}
\usepackage{graphicx}
\usepackage{amsmath}
\usepackage{amsthm}
\usepackage{booktabs}
\usepackage{algorithm}
\usepackage{algorithmic}
\usepackage[switch]{lineno}

\usepackage{microtype}      
\usepackage{xcolor}         
\usepackage{times}
\usepackage{epsfig}
\usepackage{amssymb,amsfonts}
\usepackage{multirow}
\usepackage{mathrsfs}
\usepackage{newfloat}
\usepackage{listings}

\def\BibTeX{{\rm B\kern-.05em{\sc i\kern-.025em b}\kern-.08em
    T\kern-.1667em\lower.7ex\hbox{E}\kern-.125emX}}
\begin{document}

\title{Remote Medication Status Prediction for Individuals with Parkinson's Disease using Time-series Data from Smartphones
}

\author{\IEEEauthorblockN{Weijian Li\IEEEauthorrefmark{1},
Wei Zhu\IEEEauthorrefmark{1},
E. Ray Dorsey\IEEEauthorrefmark{2,3}, and
Jiebo Luo\IEEEauthorrefmark{1}}
\IEEEauthorblockA{\IEEEauthorrefmark{1}Department of Computer Science, 
University of Rochester, Rochester, NY, USA}
\IEEEauthorblockA{\IEEEauthorrefmark{2}Center for Health + Technology and Department of Neurology, University of Rochester, Rochester, NY, USA}
\IEEEauthorblockA{Email:\IEEEauthorrefmark{1}\{wli69, wzhu15, jluo@cs.rochester.edu\}, \IEEEauthorrefmark{2}\{ray.dorsey@chet.rochester.edu\}}}

\maketitle

\begin{abstract}
Medication for neurological diseases such as the Parkinson's disease usually happens remotely away from hospitals. Such out-of-lab environments  pose challenges in collecting timely and accurate health status data. Individual differences in behavioral signals collected from wearable sensors also lead to difficulties in adopting current general machine learning analysis pipelines. To address these challenges, we present a method for predicting the medication status of  Parkinson's disease patients using the public \textit{mPower} dataset, which contains 62,182 remote multi-modal test records collected on smartphones from 487 patients. The proposed method shows promising results in predicting three medication statuses objectively: Before Medication (AUC=0.95), After Medication (AUC=0.958), and Another Time (AUC=0.976) by examining patient-wise historical records with the attention weights learned through a Transformer model. Our method provides an innovative way for personalized remote health sensing in a timely and objective fashion which could benefit a broad range of similar applications.
\end{abstract}

\vspace{-2mm}

\section{Introduction}
Parkinson's disease (PD) is the second most prevalent chronic neurodegenerative movement disorder disease in the world~\cite{rossi2018projection}. A large amount of efforts has been put into understanding~\cite{pagano2016imaging,katsuki2022cumulative}, predicting~\cite{li2020predicting,schwab2019phonemd} and providing effective treatments~\cite{ferreira2000sleep,singh2007advances} for PD. However, 
fewer studies have focused on remote PD medication monitoring. Similar to diabetes~\cite{kirk2010self}, medications for neurological disorder diseases usually need to be conducted for months or years and are usually taken remotely, e.g. in a home environment, away from hospitals. Therefore, remote medication status monitoring in a timely manner to ensure medication adherence~\cite{lakshminarayana2017using}, support dose frequency analysis~\cite{tomlinson2010systematic}, and enhance future treatment planing~\cite{zhao2021assessment} becomes an important element in disease treatment. 

To alleviate these issues, a large amount of attention has been devoted to the mobile health research~\cite{omberg2021remote,li2020predicting,wu2020personalized,schwab2019phonemd,pfister2020high}. Recently, a smartphone-based App built by Bot et al. through a research study named \textit{mPower}~\cite{bot2016mpower} provides new opportunities for remote medication status monitoring for individuals with PD. It constructs clinical-relevant PD tests on smartphones including walking test, tapping test, voice test and memory test following the standard clinical PD measurement criteria~\cite{jenkinson1997pdq}. Different from previous studies, each test is associated with a participant-reported medication point label, i.e., \textit{Immediately Before PD Medication}, \textit{Just After PD Medication} and \textit{Another Time(Other)}. The successful usage of these labeled data provides opportunities for remote medication supervision, and timely treatment plan adjustment. 

Given the large quantity of accessible smartphone test records, it becomes feasible to build machine learning models to learn feature representations and further infer medication status. Although these studies~\cite{bot2016mpower,li2020predicting} have shown encouraging progress toward disease diagnosis, little work has focused on patient medication status prediction. In particular, our goal is to validate that the same at-home multimodal sensor signals can indeed facilitate automated monitoring of fine-grained individual behaviors (e.g., medication and response to medication)~\cite{dorsey2020deep}. As a relatively well-defined behavior, medication time-point detection using machine learning models would be a natural first step or gateway toward achieving the ultimate goal.

\begin{figure*} 
\begin{center}
\includegraphics[width=0.85\linewidth]{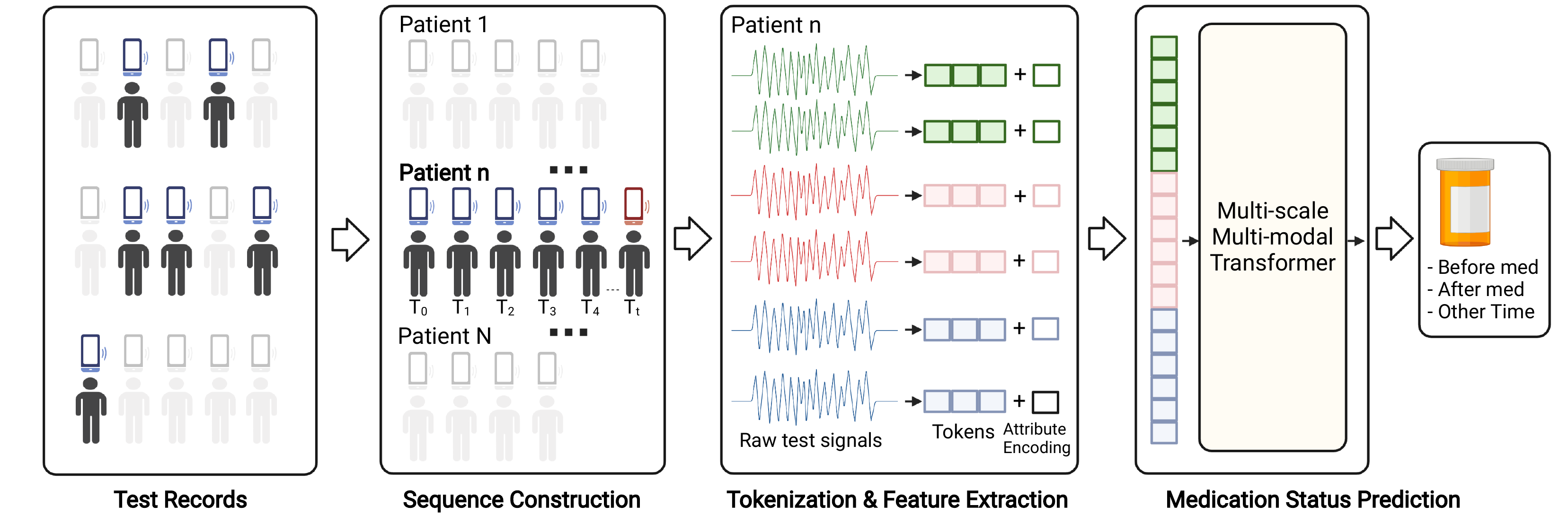}
\end{center}
   \vspace{-2mm}
   \caption{Overview of the proposed framework. Given a set of multi-modal test records from patients, we first assign the records into groups based on patient ID number and sort them chronologically. The medication status of the last record (at $T_t$) is our target. Then each record is tokenized into chunks containing subsequences of the original signal. An attribute encoding, containing the medication statuses for the previous tests, is added to the extracted features. A Transformer model takes the embedded features and predict medication status for the last record following a feature-matching mechanism. }
\label{fig:overview}
\vspace{-4mm}
\end{figure*}

To address the aforementioned issues, we introduce a framework for personalized PD medication time-point prediction with the time-series data collected in the mPower study. To be specific, we first model each patient individually by constructing patient-level records sequentially. Given several historical records whose medication status is already known, we want to predict the medication status (Before Medication, After Medication, or Another Time) for the incoming test record. A Transformer-based module then takes the tokenized and processed features as input and extracts multi-modal multi-scale feature embeddings through shuffle-and-exchange operations.

In summary, our main contributions are three-fold:
\begin{itemize}
	\item We present a framework for PD medication status prediction, i.e. 'Before', 'After', and 'Another Time', by sampling and constructing record sequences for each patient.
	
	\item Our model extracts rich time-series features through an introduced Transformer-based module with shuffle-and-merge operations, allowing multi-scale and multi-variate information exchange.
	
	\item Comprehensive quantitative evaluations on two public datasets, detailed groupwise, and individual studies, as well as attention value visualizations demonstrate the general interpretability and applicability of our method.

\end{itemize}

\section{Related Work} 
\subsection{Remote Medication Status Prediction} Significant efforts~\cite{li2020predicting,wu2020personalized,schwab2019phonemd,bohlmann2021machine,pfister2020high,rodriguez2022new,wang2022causal} have been devoted to remote health and medication status prediction in recent years. In the domain of PD, Bohlmann et al.~\cite{bohlmann2021machine} conducted an review indicating the huge potential in medication adherence prediction for PD. Schwad et al.~\cite{schwab2019phonemd} and Li et al.~\cite{li2020predicting} build deep models for PD prediction for users with smartphone based PD tests. Although these studies have shown encouraging progress towards disease diagnosis and medication adherence, limited work has focused on patient medication status prediction which is an important clinic-related out-of-lab indicator for medication adherence.

\subsection{Sequential Prediction in Healthcare}
Sequential prediction refers to a group of tasks that given the information of historical states, the model predicts a label for the current state. It has broad applications in data mining~\cite{chen2018sequential,zhang2017dynamic,joo202296}, computer vision~\cite{oh2019video,fernando2017going} and NLP~\cite{bahdanau2014neural,zhou2016attention} etc. By modeling observations sequentially, temporal relationships, as well as the individual-level attributes are naturally injected as prior knowledge into the model, allowing temporal reasoning and personalized decision-making. This idea is also widely adopted in the healthcare domain~\cite{luo2020hitanet,yin2020identifying}. Recently, Luo et al.~\cite{luo2020hitanet} introduce a time-aware attention framework for risk prediction. In their framework, a Transformer model is shown to be effective at fusing and interacting status representations among visits at different times. However, feeding all historical data into the Transformer model will lead to huge computational complexity.

\subsection{Transformer Models in Time-series Analysis} 
The Transformer~\cite{vaswani2017attention} model is originally proposed in NLP for handling long-range relationships effectively. Recently, researchers have adopted the Transformer models for processing time-series signals and reported promising results~\cite{zerveas2021transformer,li2019enhancing,akbari2021vatt,zhou2020informer}. Based on the Transformer model, Li et al.~\cite{li2019enhancing} introduce convolutional and causal self-attention for efficient time-series forecasting. The superior results indicate Transformer's strong ability not only for single modality modeling but also for multiple modalities or variables at the same time as well. Other researchers find that instance-to-instance relationship modeling boost global representations~\cite{liu2021swin,li2019patch,dosovitskiy2020image,li2020structured}. A common way to extract time-series signals representations is based on a set of predefined lengths which is insufficient to capture inner attributes such as periodic patterns due to length miss-match. Recent works~\cite{huang2021shuffle,renggli2022learning} suggest that token shuffling or merging alone provides enriched representation and reduce computational cost. However, the effective way to combine the benefits from both sides remains unexplored.

\section{Method} 
\subsection{Problem Definition} 
Given a dataset $\mathcal{X}=\{x_{1m}^{t_1},...,x_{nm}^{t_n} | n\in N,m\in M,{t_n} \in T_n\}$ contains records $x_{nm}^{t_n}$ of modality $m$, for patient $n$, at different time points ${t_n}$ in total number of time points $T_n$. For each patient's record, some modality $m$ may be missing but not all of $M$. We also have the medication status labels $y_{t_n} \in Y$, where $Y=$ $\{$\textit{Immediately Before PD Medication}, \textit{Just After PD Medication}, and \textit{Another Time(Other)}$\}$. Our goal is to predict $(T_n+1)$th record's medication status label. In other words, given the historical records of a patient, we would like to classify what the medication status $\Tilde{y}_{t_n+1}$ of the incoming record is. An illustration can be found in Figure~\ref{fig:overview}.

\subsection{Patient-level Sequence Construction}
A straightforward way of handling this classification problem is to consider all the historical records as training samples, and use the corresponding labels as supervision to train a machine learning model. Then we can use the trained model to predict the $(T_n+1)$th record in a patient-agnostic way. However, different from other classification tasks, medication status prediction is highly related to individual health and body conditions such as age, gender, disease severity, medication intake length, response to medication, etc. Besides, each person may interact with smartphones in different behavior patterns, for example, the walking speed and gait, tapping speed and strength, even in the different test environments. These disparities could easily confuse the model and lead to unsatisfied performances (please also refer to our ablative study results). 

Based on these observations, we propose to tackle this problem individually by taking patient-level knowledge into account from a query-key-matching perspective with historical records as keys and incoming records as queries. As can be seen in Figure~\ref{fig:overview}, for each patient $n \in N$, his/her records $x_{nm}^{t_n}$ are firstly grouped based on patient ID and are sorted based on time. After processing, record embeddings $v_{nm}^{t_n} \in \mathbb{R}^d$ at $t_n \in \{1,...,T_n\}$ are merged with the corresponding medication status embeddings $v_s \in \mathbb{R}^d$. Encoding details can be found in the next subsection. 

Notice that patients may have conducted different number of tests ($T_n$ is different). Directly computing query-key featuring matching will bring significant computation overhead especially for patients who conduct more tests than the others ($T_n$ is larger). To overcome this issue, for each patient, we sample $K$ records randomly from their history records at time points $T'_n$ instead of feeding all records into the model. By adopting this few-shot training strategy, our model is not only more robust in tackling the overfitting problem, but also better handling patients with sparse records, i.e. small or different amount of test records in different modalities, which is commonly observed in real world.

\begin{table}[!t]
\small
  \centering
    \caption{Statistics of the dataset after preprocessing. $\#$ represents that the numbers are counts. $\%$ represents that the numbers are in percentage. Notice that each patient may have multiple test records in different medication statuses, but the same patient along with his/her test records only exists in either training or testing set to avoid data leak.}
  \resizebox{1\linewidth}{!}{
      \begin{tabular}{l|ccc}
        \toprule
                \textbf{Properties} & \textbf{Another Time} & \textbf{Before Med.} & \textbf{After Med.}\\
        \midrule
                Patient gender: Female (\#) \& (\%)  & 163 (41.9) & 118 (41.4) & 125 (40.3) \\
                Patient age: Mean \& STD  & 63.1 (7.7) & 63.4 (7.2) & 63.3 (7.6) \\
        \midrule
                Records all: (\#) \& (\% over classes) & 31,493 (50.6) & 15,110 (24.3) & 15,579 (25.1) \\
                Records all: Tap (\#) \& (\% over classes) & 17,699 (49.9) & 8,477 (24.0) & 9,241 (26.1) \\
                Records all: Walk (\#) \& (\% over classes) & 10,749 (51.3) & 5,308 (25.4) & 4,865 (23.3) \\
                Records all: Mem. (\#) \& (\% over classes) & 3,045 (52.1) & 1,325 (22.7) & 1,473 (25.2) \\
                Records all: Tap Walk Mem. (\% over records) & 56.2/34.2/9.6 & 56.1/35.2/8.7 & 59.3/31.2/9.5 \\
                Records per patient: all (\#) (Mean \& STD) & 80.9 (123.1) & 52.9 (90.8) & 50.2 (73.0) \\
                Records per patient: Tap (\#) (Mean \& STD) & 45.5 (61.5) & 29.7 (46.0) & 29.8 (42.4) \\
                Records per patient: Walk (\#) (Mean \& STD) & 27.6 (51.9) & 18.6 (39.7) & 15.6 (27.8) \\
                Records per patient: Mem (\#) (Mean \& STD) & 7.8 (21.6) & 4.6 (12.5) & 4.8 (16.8) \\
        \bottomrule
      \end{tabular} 
  }
\label{tab:stat}
\vspace{-4mm}
\end{table}

\subsection{Tokenization and Attribute Encoding} After sampling and constructing record sequence for each patient 
our model takes a sequence of raw time-series signals at new sampled time points as input:
\begin{equation}
	\label{eq:trans_r}
	X_n=\{x_{n1}^{t_n},...,x_{nm}^{t_n}|m\in M,{t_n} \in T'_n\},
\end{equation}
where each time series $x_{nm}^{t_n} \in \mathbb{R}^{L\times 3}$ consists of 3 channels representing x, y, and z dimensional accelerometer readings. Similar to previous studies~\cite{dosovitskiy2020image,akbari2021vatt}, we flatten and chunk a time-series signal into $P$ 1D segments $x'\in \mathbb{R}^{P \times (S\cdot 3)}$, where $S$ is a predefined length of the 1D segments, and $P=\frac{L}{S}$ is the number of segments obtained. A linear layer is then used to project the chunked segments into feature vector representations:
\begin{equation}
	\label{eq:trans_v}
	{v_{nm}^{t_n}}^T=W^T \cdot x_{nm}^{t_n} + b,
\end{equation}
where $W \in \mathbb{R}^{(S\cdot 3)\times d}$, $b \in \mathbb{R}^{d}$, $v_{nm}^{t_n} \in \mathbb{R}^{P\times d}$. After applying the same operation on all time points, we obtain a set of $T'_n$ tokenized and embedded time-series signals. 

We follow previous works~\cite{dosovitskiy2020image,akbari2021vatt,vaswani2017attention} to use learnable embeddings to encode different elements:
\begin{enumerate}
  \item Positional Encodings $v_p \in \mathbb{R}^{P \times d}$: each one corresponds to the original position of the chunked segment in the raw time-series signal to fill in the missing positional information.
  \item Time Encodings $v_t \in \mathbb{R}^{24 \times d}$: each one represents the hour the test is conducted. This encoding is used to capture medication time and human body condition periodic patterns.
  \item Modality Encodings $v_o \in \mathbb{R}^{M \times d}$: each one indicates the modality that the time-series corresponds to. Another motivation of adopting this encoding is to learn cross-patient modality-modality relationships.
  \item Status Encodings $v_s \in \mathbb{R}^{3 \times d}$: each one corresponds to the status we want to predict. As mentioned in the previous subsection, we leverage this encoding to build a personalized medication status prediction model based on a query-key matching strategy. Used only for the historical records.
\end{enumerate}

Encodings, $v_p'$, $v_t'$, $v_o'$, $v_s'$, are extracted from $v_p$, $v_t$, $v_o$, $v_s$ based on the index in the corresponding input information, i.e. the position of the segment, the time in 24 hours, the modality the test is in, and the status when the test is performed. Feature embeddings is then updated by merging with the aforementioned encoddings:
\begin{equation}
	\label{eq:mask_merging}
	v_{nm}^{t_n} = v_{nm}^{t_n} + v_p' + v_t' + v_o' + v_s' \textrm{(if history)},
\end{equation}

\begin{figure}[t]
\begin{center}
\includegraphics[width=0.8\linewidth]{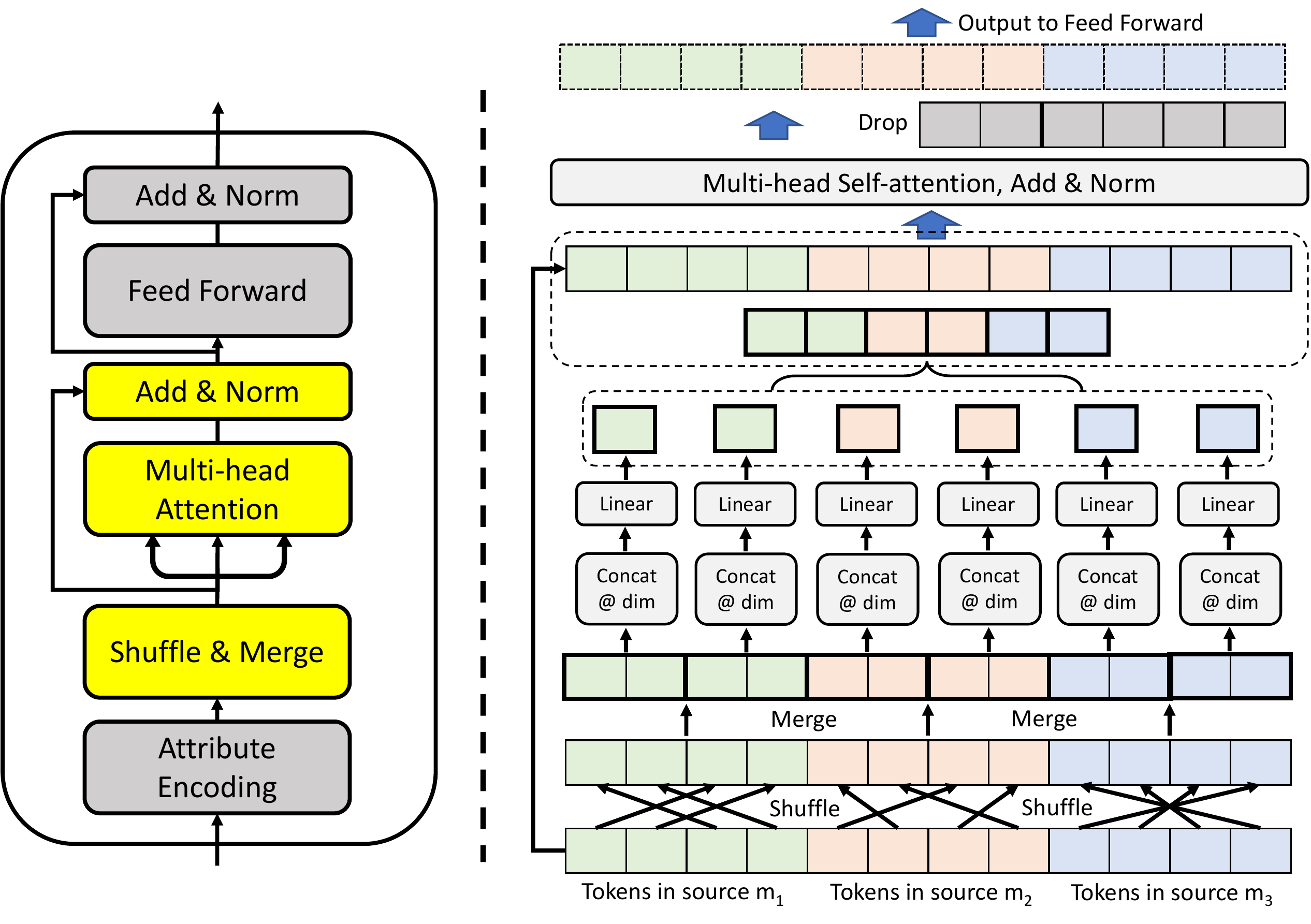}
\end{center}
\vspace{-2mm}
  \caption{Overview of the proposed Transformer module. Each of the time-series signals is first tokenized and projected into embeddings. Token shuffling and merging are conducted to aggregate remote tokens.}
\label{fig:transformer}
\vspace{-4mm}
\end{figure}
\subsection{Multi-scale Transformer} 
The Transformer Model~\cite{vaswani2017attention} is an encoder-decoder structure where each encoder and decoder consists of multiple layers of the attention blocks. However, due to the nature of time-series signals, feature scale and the length of time series signals are crucial in determining the representation of the patient status. The pre-defined segment length of each sequence sent into transformer limits this multi-scale representation. Directly combining neighboring sub-sequences as new representations may generate enlarged-scale information but does not consider long-term larger-scale representations. Inspired by the ShuffleNet model~\cite{zhang2018shufflenet}, we present a shuffle-and-merge approach to extract multi-scale time-series representations. Each layer of the Shuffle Encoder mainly contains three parts: 1) token-shuffling, 2) token-merging, and 3) multi-head self-attention and add \& Norm. Different from the ShuffleNet model~\cite{zhang2018shufflenet}, our method focuses on constructing multi-scale representations by shuffling time-series sequence in the temporal dimension, and proposes a merge-and-drop operation that naturally fits the need in our Transformer structure.

Firstly, token-shuffling operation shuffles the tokenized and attribute aggregated tokens from the previous module according to a random order: 
\begin{equation}
	\label{eq:token-shuffling}
	{v_{nm}^{t_n}} = \textrm{Shuffle}(\{v_{nm1}^{t_n},v_{nm2}^{t_n},...,v_{nmp}^{t_n}\}) 
\end{equation}
By doing this, we reallocate each segment leading to a set of neighbor-changed tokens. This makes previously remote tokens to become closer and previously neighboring tokens to be further apart. 

Then token-merging operation takes the shuffled tokens and groups each pair of the neighboring tokens together to become a combined feature representation by concatenation and mapping:
\begin{equation}
	\label{eq:token-linear}
	{v_{nm}^{t_n}}' = \textrm{Linear}\{\textrm{Concat}([{v_{nmk_i}^{t_n}},{v_{nmk_{i+1}}^{t_n}}])\}
\end{equation}
where $\{k_i\}$ represent the new indices for the shuffled tokens. 
Notice that shuffling and merging operations are conducted within each source in this paper. With the help of these two operations, we obtain an additional group of tokens where each of them contains a combined representation from two previous tokens. We call this \textit{the second-order tokens}. Then the multihead self-attention, add$\&$norm same as the original Transformer model is applied to the first-order and the second-order tokens from all modalities and at all time points:
\begin{equation}
	\label{eq:selfatt}
	\textrm{Att\&AN}(\{v_{n1}^{t_n},...,v_{nm}^{t_n}, {v_{n1}^{t_n}}',...,{v_{nm}^{t_n}}'|m\in M,{t_n} \in T'_n\}),
\end{equation}
Then merged tokens are dropped. This completes one layer of the proposed Shuffle-Encoder. We reassign the shuffled tokens as the previous tokens and send them to the rest operations:
\begin{equation}
	\label{eq:token-merging}
	v_{n} = \textrm{Add\&Norm}(\textrm{FeedForward}(v_{nm}^{t_n}))
\end{equation}
The final prediction is obtained by computing a global average pooling and mapping on the output tokens of the last layer:
\begin{equation}
	\label{eq:avgpool}
	y_{n} = \textrm{MLP}\{\textrm{AvgPool}(v_{n})\}
\end{equation}

\subsection{Training}
By examining the class imbalance problem, we train our model with a weighted cross-entropy loss where the weights are assigned to each class based on the class distribution in the batch:
\begin{equation}
    \label{eq:loss}
    \mathcal{L}=-\frac{1}{N}\sum_{i=1}^{N} \sum_{c=1}^{C}w_{c}y_{ic}log(\frac{e^{\hat{y_{ic}}}}{\sum_{j=1}^{\hat{y_{jc}}}})
\end{equation}
where $w_c$ are classwise weights. Their values are inversely proportional to the classwise sample ratios in the batch.

\section{Experiments}

\subsection{Dataset Description} 
The \textit{mPower} study~\cite{bot2016mpower}\footnote{\url{https://www.synapse.org/\#!Synapse:syn4993293/wiki/247859}} is a clinical observational study for Parkinson's disease through an iPhone App interface. The goal of this study is to provide an opportunity to collect and examine high-resolution and more frequent out-of-lab quantitative assessments~\cite{bot2016mpower}. Participants involved in the study are asked to select and perform the tests at most 3 times at any time of the day on their smartphones. In this study, we adopt three kinds of test records: Tapping Test, Walking Test, and Memory Test. Details of these three tests can be found in supplementary material. Each test also provides a selection (label) on the participant's medication status, and the label choices are: \textit{Immediately before Parkinson medication}, \textit{Just after Parkinson medication (at your best)}, and \textit{Another time(Other)}. We keep all three labels as our target variables for comprehensive prediction of patient's daily medication status. To further evaluate our model, we conduct additional evaluation on the PhysioNet-2019 dataset~\cite{reyna2019early}, details can be found in supplementary materials.

\subsection{Data Construction} \label{subsection:preprocess}
We follow previous studies on the mPower dataset~\cite{li2020predicting,schwab2019phonemd} to preprocess the mPower dataset. Since patients are free to choose when and what types of tests to conduct, this leads to sparsely observed and distributed tests records across time. To conduct multimodal analysis, we temporally synchronize test results across different modalities based on a predefined time-window to obtain multimodal observations at unified time points. Other dataset collection details can be found in~\cite{bot2016mpower}.

\noindent\textbf{Record Construction} 
For the tapping and walking tests, smartphone accelerometer readings that contain four-dimensional $(time, x, y, z)$ sequences are extracted as the raw time-series signals. We then apply the high-pass filters~\cite{badawy2018automated} to the time-series sequences to remove the gravitational component. For the memory tests, button-tapping sequences as well as the corresponding target button sequences are extracted to form the time-series sequences. Each touch is also attached with a game score that overall gives us a four-dimensional representation for each touch: $(time, actual, target, score)$. The maximum lengths of walking, tapping, and memory series are set to be $L_{walk}=L_{tap}=1024$, and $L_{memory}=32$. Zeros will be appended to the record if the record length is smaller than $L$.

\begin{table}[t]
  \centering
    \caption{Evaluation results on the mPower dataset for three statuses under 5-fold cross-validation. All models leverage the constructed 5 record sequences as input. $\pm$ represents the mean value and the standard deviations of the five folds.} 
    \resizebox{0.9\linewidth}{!}{
      \begin{tabular}{lccc}
        \toprule 
            \textbf{Method} & \textbf{Accuracy} & \textbf{F1 Score} & \textbf{AUC} \\
        \cmidrule{1-4}
        \multicolumn{4}{c}{\textbf{Quantitative Evaluation}} \\
        \cmidrule{1-4}
                Random Guess & 0.397$\pm$0.025  & 0.335$\pm$0.019 & 0.507$\pm$0.016\\
                XGBoost & 0.817$\pm$0.017& 0.792$\pm$0.021& 0.935$\pm$0.023\\
                MLP & 0.842$\pm$0.016 & 0.814$\pm$0.033 & 0.944$\pm$0.006 \\
                TCN & 0.858$\pm$0.004 & 0.838$\pm$0.013 & 0.948$\pm$0.004 \\
                Bi-LSTM & 0.850$\pm$0.018 & 0.825$\pm$0.027 & 0.944$\pm$0.007 \\
                PD & 0.871$\pm$0.012 & 0.859$\pm$0.021 & 0.960$\pm$0.006 \\
                ConvSelfAttn & 0.873$\pm$0.017 & 0.866$\pm$0.027 & 0.958$\pm$0.006 \\
                VATT & 0.896$\pm$0.016 & 0.885$\pm$0.027 & 0.960$\pm$0.008 \\
                \textbf{Ours} & \textbf{0.918$\pm$0.016} & \textbf{0.901$\pm$0.029} & \textbf{0.969$\pm$0.006} \\
        \bottomrule
      \end{tabular}
      \label{tab:quantitative_mpower}
      }
\vspace{-4mm}
\end{table}

\noindent\textbf{Record Synchronization}
In this part, we would like to temporally synchronize records for each participant based on timestamps to construct multmodal observations. First, test records with the same participant ID and medication status are grouped together. For each group, we sort the records chronologically and further merge the records within 30 minutes into subgroups without duplication. Each subgroup now contains one or more test records that may come from different test modalities, i.e. walking test, tapping test and memory test. These subgroups are considered as our multimodal observations at unified time points. Last, we sort the subgroups based on each group's average observation time of the three modalities to give us the constructed sequential multimodal records for each patient.

\noindent\textbf{Other processing} 
We adopt a similar approach as previous studies~\cite{schwab2019phonemd,prince2018multi,li2020predicting} to extract qualified participants. Since we are more interested in predicting PD medication status, we filter out users labeled as \textit{Non-PD}, and only test records that are labeled with the three target statuses. To ensure enough samples to construct sequential representation, participants who perform fewer than 6 tests in total are also not included in the study.

\subsection{Comparing Methods} 
To investigate the effectiveness of the proposed method, we examine the model performance on the mPower dataset.
We first try random guessing one of the three statuses which is considered as our baseline method. Then different types of learning-based methods are adopted including: \textbf{(1)} the classic methods: XGBoost~\cite{chen2016xgboost}, MLP~\cite{gardner1998artificial}, TCN~\cite{lea2017temporal}, and BiLSTM~\cite{graves2013hybrid}, \textbf{(2)} Transformer based methods in the time-series analysis domain: ConvSelfAttn~\cite{li2019enhancing} and VATT~\cite{akbari2021vatt}, \textbf{(3)} a PD prediction method(PD) with the mPower dataset~\cite{li2020predicting}.

\section{Results}
After pre-processing (sec. \ref{subsection:preprocess}), there are 487 individuals with PD included in our study with test records conducted between March 09, 2015 and September 04, 2015. Among them, 189 (40.1\%) are female, age ranges from [45, 86] with a mean of 63.2. 163 (41.9\%), 118 (41.4\%), and 125 (40.3\%) of them have the last test record labeled in \textit{Another Time}, \textit{Before Medication}, and \textit{After Medication} respectively. In total, there are 62,182 test records included and 35,417 (56.9\%) are tapping tests, 20,922 (33.7\%) are walking tests, and 5,843 (9.4\%) are memory tests. Details of the dataset information can be found in Table~\ref{tab:stat}.

\begin{figure}[t]
\begin{center}
\includegraphics[width=0.9\linewidth]{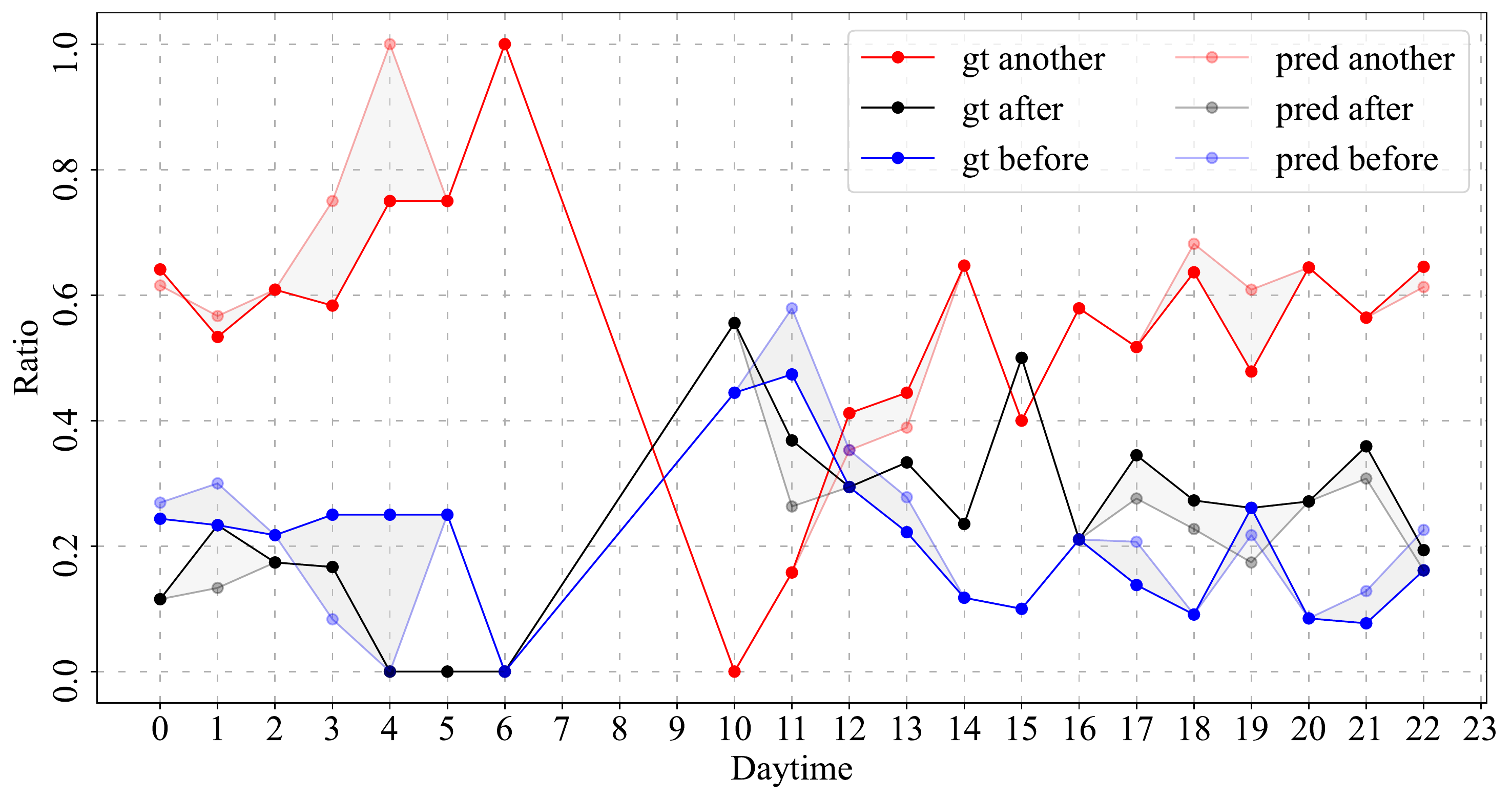}
\end{center}
   \vspace{-2mm}
  \caption{Ratio distribution of three medication statuses at different times of the day. Ratio is computed by the number of records in the specified medication status divided by the total number of records at the time. Gray shadow: the difference between groundtruth and prediction.}
\label{fig:time}
   \vspace{-6mm}
\end{figure}

\begin{figure*} 
\begin{center}
\includegraphics[width=0.9\linewidth]{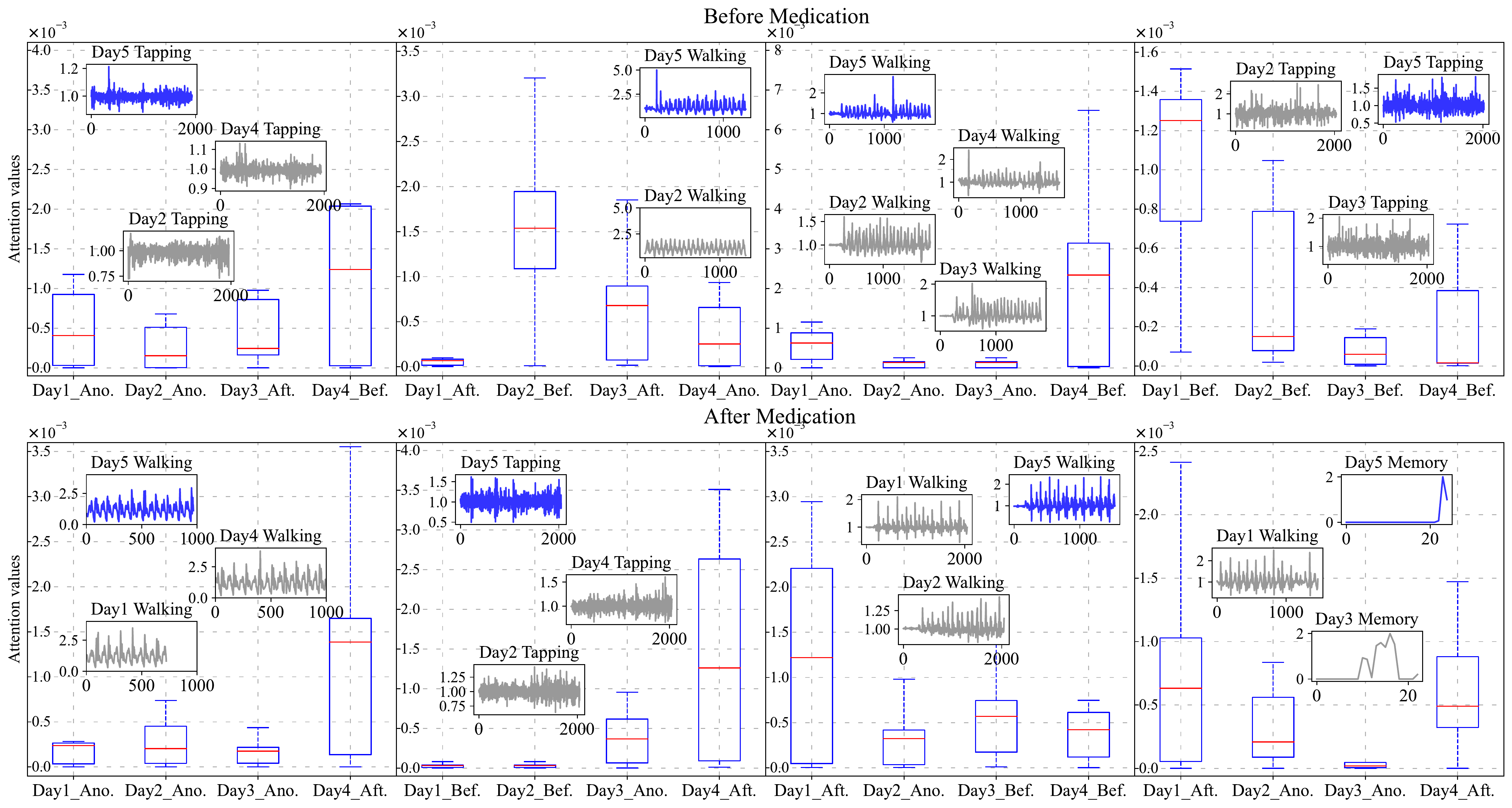}
\end{center}
   \vspace{-2mm}
  \caption{Bar plot for attention values for all segments from three modalities at each historical record using the fifth day's test. Blue time-series: the sample signal at the current incoming test. Grey time-series: the sample signal at previous days. Best viewed by zoom-in on screen. }
\label{fig:att_boxplot}
\vspace{-4mm}
\end{figure*}

\subsection{Performance Evaluation}
\noindent{\bf K-fold Validation}
We conducted a K-fold validation (K=5) to validate our method. In detail, the dataset was first shuffled and was split into five subsets with similar test record sizes and class distributions according to the participant's user ID. At each time, four folds were selected as training data, and the remaining one fold was used as testing data (no participant overlaps between training and testing). After obtaining the evaluation results we chose another fold as testing data and the rest four folds as training data. This process continued until all 5 folds were evaluated. We examined our method on various evaluation metrics and found promising results. For example, our method achieves an average 0.969 AUC score, and an average 0.901 F1 score for the three target medication variables. It outperformed the standard machine learning approaches (directly predicting medication status without considering patient identity and record sequences) (Table~\ref{tab:quantitative_mpower}).

\noindent{\bf Comparison to Other Methods} We find that almost all models achieve a high F1 score of $0.792$ or higher as is shown in Table~\ref{tab:quantitative_mpower}. This indicates the effectiveness of the proposed sequential modeling strategy in the medication status prediction task. A closer look at the table, we find that almost all neural network based models perform better than the non-neural network model, i.e. XGBoost. We believe higher model capacity would lead to better results in our task. Comparing to non-Transformer models, i.e. MLP, TCN and Bi-LSTM, we observe that Transformer-based models achieve higher performance. Recall that ConvSelfAttn model leverages a convolution filter to aggregate segments in local neighborhoods. This operation brings segments together and is supposed to enable locality awareness similar to our approach. However, simply merging segments unavoidably lead to more complicated representation and lost the representation by original individual segments. This indicates that smaller segments are actually important in the Transformer models.  We also notice that our method achieves better performance compared to a recent domain specific method for PD classification~\cite{li2020predicting}. We find that our model outperforms all previous methods including the lately proposed Transformer-based method ConvSelfAttn and VATT. In particular, our method achieves 0.918 and 0.901 in accuracy and F1 score that exceed VATT by almost $2\%$.

\subsection{Overall Medication Pattern Visualization}
Patient test record patterns could reflect test time preferences which could be used to improve remote health assessment designs for better enrollment and experience (e.g. test length per time). As shown in Figure~\ref{fig:time}, we draw line plots on the ground-truth medication statuses as well as the predicted medication statuses across 24 hours based on all the records from participants in the testing set. For each hour, the ratios of patients for the three medication statuses are plotted with three different colors, respectively. First, we observe that most patients prefer to take a test after their medication around 10 a.m. and 15 p.m., and are less likely to do so in the early mornings or late nights. Participants prefer to conduct tests on these timeframes and label the test records as Another Time. By looking at the prediction results of our model, we find that they align with the ground-truth ratios well with slight drifts in a local range (0.04$\pm0.05$). Recall that our method uses a small number of historical records (4) for each participant to construct a record sequence to make a prediction on the incoming test record. This few-shot design would be particularly useful in reducing each participant's efforts in the data collection process in related human-subject studies.

\subsection{Prediction Interpretation}
Next, we demonstrate the model's ability on explaining the prediction decisions quantitatively and qualitatively with the attention value. Eight box plots are plotted in Figure~\ref{fig:att_boxplot}. Each of them corresponds to an incoming record from a different participant labeled with either Before Medication or After Medication. Inside each plot, all the attention values on the segments (see Methods) of four historical test records are drawn as blue boxes. The attention computation is conducted during the self-attention operation in the Transformer model. Higher attention values represent higher weights assigned to records that contain similar features. As can be seen from this figure, a stronger relationship is discovered between records with the same medication status (e.g. Day 4, Day 2, Day 4, and Day 1 in the first row; Day 4, Day 4, Day 1, and Day 1 in the second row), weaker relationships are assigned to records with different medication status. Records with the same medication status usually share similar time-series patterns (e.g. Day 5 and Day 4 in the top left corner), and records with different medication statuses are shown different patterns (e.g. Day 5 and Day 2). When there are multiple records with the same medication status, the system tends to assign higher weights to records with closer patterns (e.g. Day 5 and Day 2 in the upper right corner), or assign lower weights if a clear disagreement is observed (e.g. Day 5 and Day 3 in the lower right).

\section{Conclusion} 
In summary, we present a novel interpretable deep learning based framework for PD medication status prediction. The proposed framework models patient test records as temporal sequences and predicting current status by querying historical records. This mechanism largely increased prediction performance by modeling individual differences. The sequential record construction process requires fewer samples to achieve strong performance suitable for a variety of medical problems. Comprehensive experiments on a large public PD study dataset demonstrate the stability and explainability of the proposed methods in PD medication status prediction. 

\section{Acknowledgement} Research reported in this publication was supported by the National Institute Of Neurological Disorders And Stroke of the National Institutes of Health under Award Number P50NS108676. The content is solely the responsibility of the authors and does not necessarily represent the official views of the National Institutes of Health.

\section{Supplementary Material}
\subsection{Implementation Details}
Since the XGBoost and MLP models are not designed for handle sequential data, we flatten the sequences, and concatenate them together to feed into the models. For TCN and BiLSTM models, we concatenate modality tests at each time point and feed the record as a sequence into the model. We use the same tokenized records as inputs for LogSparse and VATT as we use in our model. All Transformer-based models contain 6 layers and each self-attention operation has 8 heads. AdamW optimizer is used with a learning rate equal to 1e-5 to train our model. The proposed model is implemented in PyTorch and trained on a single NVIDIA Geforce GTX 1080Ti GPU with batch size equal to 2.

\subsection{Comparison on PhysioNet-2019}
To further evaluate our model, we conduct additional evaluation on the PhysioNet-2019 dataset~\cite{reyna2019early}. \textit{PhysioNet-2019} dataset is collected from ICU patients from three hospital systems and is used as a challenge dataset for Sepsis early detection. We follow the same preprocessing steps in a previous study~\cite{kidger2020neural} for this dataset where each patient's first 72 hours of stay is extracted as time-series signals. 34 time-dependent physiological features such as respiration rate, creatinine concentration in the blood as well as 5 demographic features are used as multi-source features. The dataset is partitioned into train/val/test following the 70\%/10\%/20\% ratio. Other dataset details can be found in previous works~\cite{reyna2019early,kidger2020neural}. 

Different from the mPower dataset, it consists of multi-variate irregular time-series where the multi-source time-series signals are physiological measurements through time but potentially missing which is the key challenge in this dataset. Comparing to previous methods on this dataset, we find that our model is able to outperform the strong ODE/CDE-based methods (GRU-ODE~\cite{de2019gru}, ODE-RNN \cite{chen2018neural}, Neural-CDE \cite{graves2013hybrid}) which construct dynamic state-wise transitions in latent space with ODEs under both OI and No OI criterions. Though we did not follow the same strategy as the these methods, our model is able to handle the irregularly observed time-series data and provide even better predictions. That said, our approach is parallel to the ODE/CDE based models. Aggregating ODE methods into our model may further increase the performance. When removing the observation mask, the performance of our model drops by a large margin of 5\%. We infer that observational status is an important part to the success of self-attention computation. Without these observational attributes, the self-attention computations are hard to focus on the key segments especially for time-series that contain too much noise in the signal.

\begin{table}[t!]
  \centering
    \caption{Additional evaluation on the PhysioNet-2019 dataset. We run the model 5 times following previous approaches. $\pm$ represents the mean value and the standard deviations of the five runs. OI: include observation mask. No OI: without observation mask.} 
    \resizebox{0.8\linewidth}{!}{
      \begin{tabular}{lcc}
        \toprule 
                \multirow{2}{*}{\textbf{Method}} & \multicolumn{2}{c}{\textbf{Test AUC}} \\
        \cmidrule(lr){2-3}
                & \textbf{OI}  & \textbf{No OI} \\
        \cmidrule{1-3}
                GRU-ODE & 0.852$\pm$0.010 & 0.771$\pm$0.024 \\
                GRU-$\Delta t$ & 0.878$\pm$0.006& 0.840$\pm$0.007\\
                GRU-D &  0.871$\pm$0.022 & 0.850$\pm$0.013 \\
                ODE-RNN & 0.874$\pm$0.016 & 0.833$\pm$0.020 \\
                Neural-CDE & 0.880$\pm$0.006 & 0.776$\pm$0.009 \\
                \textbf{Ours} & \textbf{0.909$\pm$0.008} & \textbf{0.852$\pm$0.011}\\
        \bottomrule
      \end{tabular}
      \label{tab:quantitative_physionet}
      }
\end{table}

\begin{table}[t!]
  \centering
    \caption{Ablative Study on the mPower dataset.}
    \resizebox{0.9\linewidth}{!}{
      \begin{tabular}{lccc}
        \toprule 
                \textbf{Ablatives} & \textbf{Accuracy} & \textbf{F1 Score} & \textbf{AUC} \\
        \cmidrule{1-4}
        \multicolumn{4}{c}{\textbf{Effect of Patient Sequence Modeling}} \\
        \cmidrule{1-4}
                $\textrm{No Seq. Modeling}$  & 0.398 & 0.373 & 0.584 \\
                $\textrm{Seq. Modeling}$  & \textbf{0.918} & \textbf{0.901} & \textbf{0.969} \\
        \cmidrule{1-4}
        \multicolumn{4}{c}{\textbf{On the merging length}} \\
        \cmidrule{1-4}
                $G=1$  & 0.903 & 0.887 & 0.958 \\
                $G=2$  & \textbf{0.918} & \textbf{0.901} & \textbf{0.969} \\
                $G=4$  & 0.911 & 0.896 & 0.961 \\
        \cmidrule{1-4}
        \multicolumn{4}{c}{\textbf{On the attribute encodings}} \\
        \cmidrule{1-4}
                $\textrm{No Encodings}$ & 0.505 & 0.325 & 0.565 \\
                $\textrm{+ Status Encodings}$  & 0.797 & 0.771 & 0.933 \\
                $\textrm{+ Positional Encodings}$  & 0.843 & 0.833 & 0.946 \\
                $\textrm{+ Modality Encodings}$  & 0.893 & 0.876 & 0.958 \\
                $\textrm{+ Time Encodings (all)}$ & \textbf{0.918} & \textbf{0.901} & \textbf{0.969} \\
        \cmidrule{1-4}
        \multicolumn{4}{c}{\textbf{On the token shuffle-merge}} \\
        \cmidrule{1-4}
                $\textrm{No shuffle-merge}$ & 0.887 & 0.866 & 0.953 \\
                $\textrm{+ Shuffle-merge}$   & \textbf{0.918} & \textbf{0.901} & \textbf{0.969}\\
        \bottomrule
      \end{tabular}
      \label{tab:ablative}
      }
\vspace{-2mm}
\end{table}

\subsection{Ablative Studies}
In this subsection, we examine the performance of the proposed approach by conducting ablative studies on the mPower dataset.

\noindent{\bf Effect of Sequence Modeling.}
We first examine the effect of sequence modeling. As we can see in Table~\ref{tab:ablative}, sequence modeling significantly enables the learning process. We consider this is due to the differences among different smartphone usage preferences, test environments as well as motion patterns.

\noindent{\bf Choices on attribute encodings.}
We study different choices and combinations on the attribute encodings. As is depicted in Table~\ref{tab:ablative}, our model achieves a significant improvement when Status Encodings are aggregated into the model. Recall that, our patient-level sequential modeling transforms the original classification task as a patient sequence prediction task. This helps our model to focus on individual mobile phone usage patterns which is a strong prior knowledge that assist decision making considering the potential noisy environments when experiments are conducted. By integrating each of the encodings into our model, we observe consistent performance improvements. This indicates the benefits of attribute encodings in our task. The best performance is achieved when all attribute encodings are aggregated into the feature.

\noindent{\bf Choices on the number of tokens in token-merging.}
We analyze the effect of the number of tokens in token-merging. As can be seen from Table~\ref{tab:ablative}, when $G=1$, the model becomes a traditional transformer encoder. When $G=2$, we aggregate each pair of the tokens together and achieve the best performance. When $G=4$, we find the performance slightly drops. We consider this is because of a too long token which diminish the smaller segment information. A better choice of $G$ is important and should be experimented based on the real-world task.

\noindent{\bf Effect of token shuffling operation.}
Token shuffling and merge operation is introduced in this paper to facilitate multi-scale information aggregation from distant and non-neighboring tokens. As we can see from Table~\ref{tab:ablative}, when introducing this module into our model, the overall performance gain a large improvement, for example, from 0.866 to 0.901 in F1 Score. This indicates the importance of modeling long-distance multi-scale relations between tokens.

\begin{figure} 
\begin{center}
\includegraphics[width=0.9\linewidth]{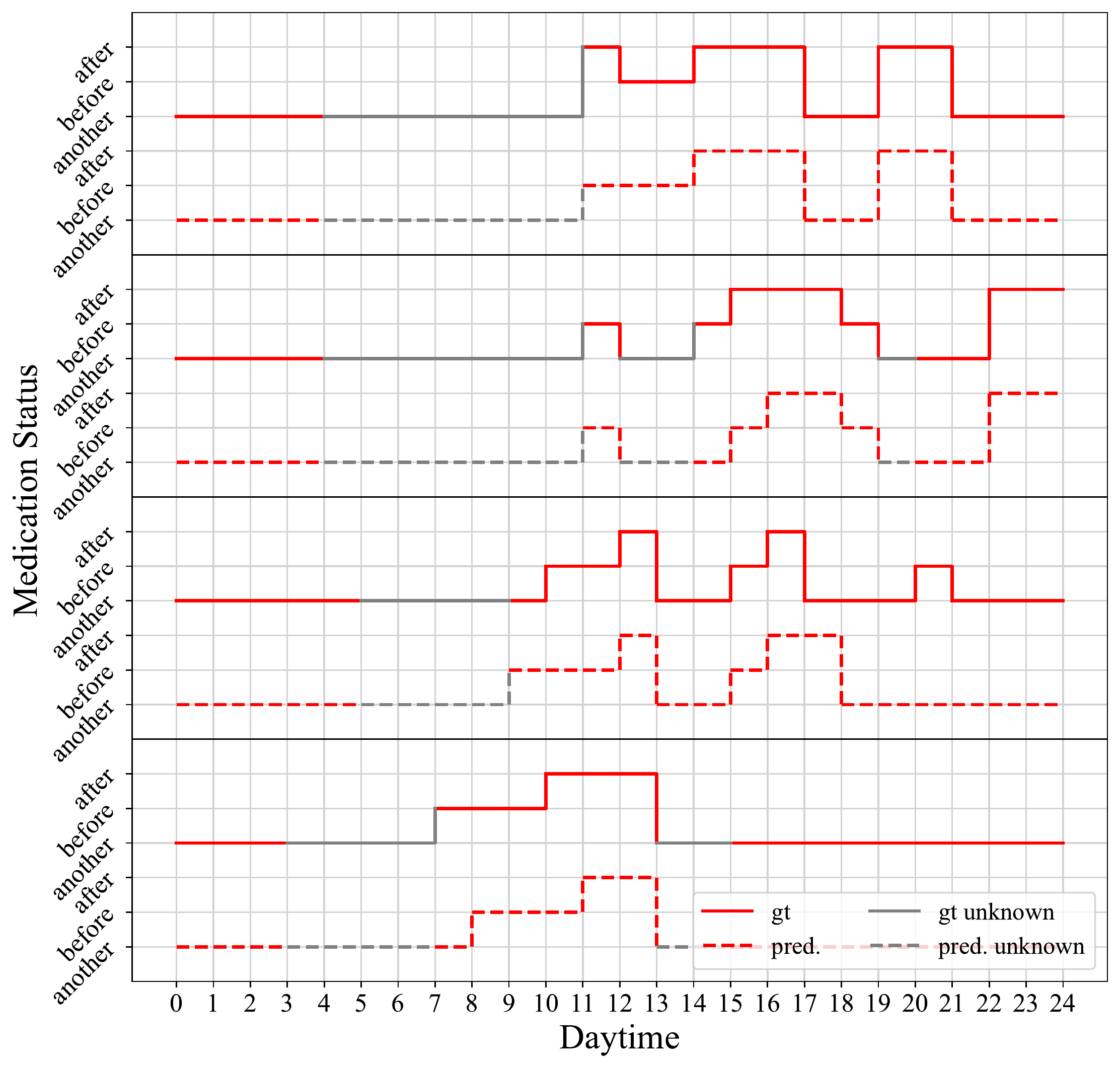}
\end{center}
   \vspace{-2mm}
  \caption{Status patterns visualization for four participants. Each row: each participant's temporal transitions of medication statuses.}
  \vspace{-4mm}
\label{fig:individual}
\end{figure}

\subsection{Individual Medication Pattern Visualization}
Furthermore, we examine the statistical medication status patterns for four individual participants (Figure~\ref{fig:individual}). Different from previous experiment settings, here we predict participant's entire records using his/her 4 oldest records as the historical records. For example, suppose a participant has 10 records in total. The last 6 records (6 of 10) are our predict targets. Each one of the six is predicted by making the 5-tuples where 4 of the tuples are the first 4 records (4 of 10), and 1 of them is the target record (1 of 6). This is done with the same trained model used in all other experiments. Similarly to Figure~\ref{fig:time}, we compute the ratios of the three medication statuses at each time point. However, instead of showing the ratios of the three medication statuses, we show the status with the highest ratio at each time point,  which indicates the most likely status at that time. This provides an overall impression of the statistical medication status patterns for each patient. Overall, our method shows a good ability in capturing the daily medication patterns including the alignment of the major transitions.

\textit{Tapping Test} is designed to measure dexterity and speed. Participants are instructed to place their smartphone on a flat surface and use two fingers on the same hand to tap the two buttons shown on the screen alternatively for 20 seconds. This test records smartphone sensor signals such as accelerometer readings and actual tapping sequences as time-series sequences.

\textit{Walking Test} is designed to evaluate gait and balance. During the test, participants need to place their smartphone in the pocket then follow the instructions to walk 20 steps in a straight line, stand still for 30 seconds, then walk 20 steps back. This test records accelerometer and gyroscope readings as time-series sequences. 

\textit{Memory Test} is designed to examine short-term spatial memory. During the test, participant are asked to sit and pay attention to their smartphone screen. A grid of flower figures will be shown on the screen and illuminated one at a time in a random order. After a sequence of illuminations, participants are asked to repeat the sequence by tapping the corresponding flowers in the grid. This test records groundtruth and actual tapping sequences, as well as an evaluation score given by the system. 

\subsection{Group-wise Results}
\begin{figure}[tb]
\minipage{0.24\textwidth}
  \includegraphics[width=\linewidth]{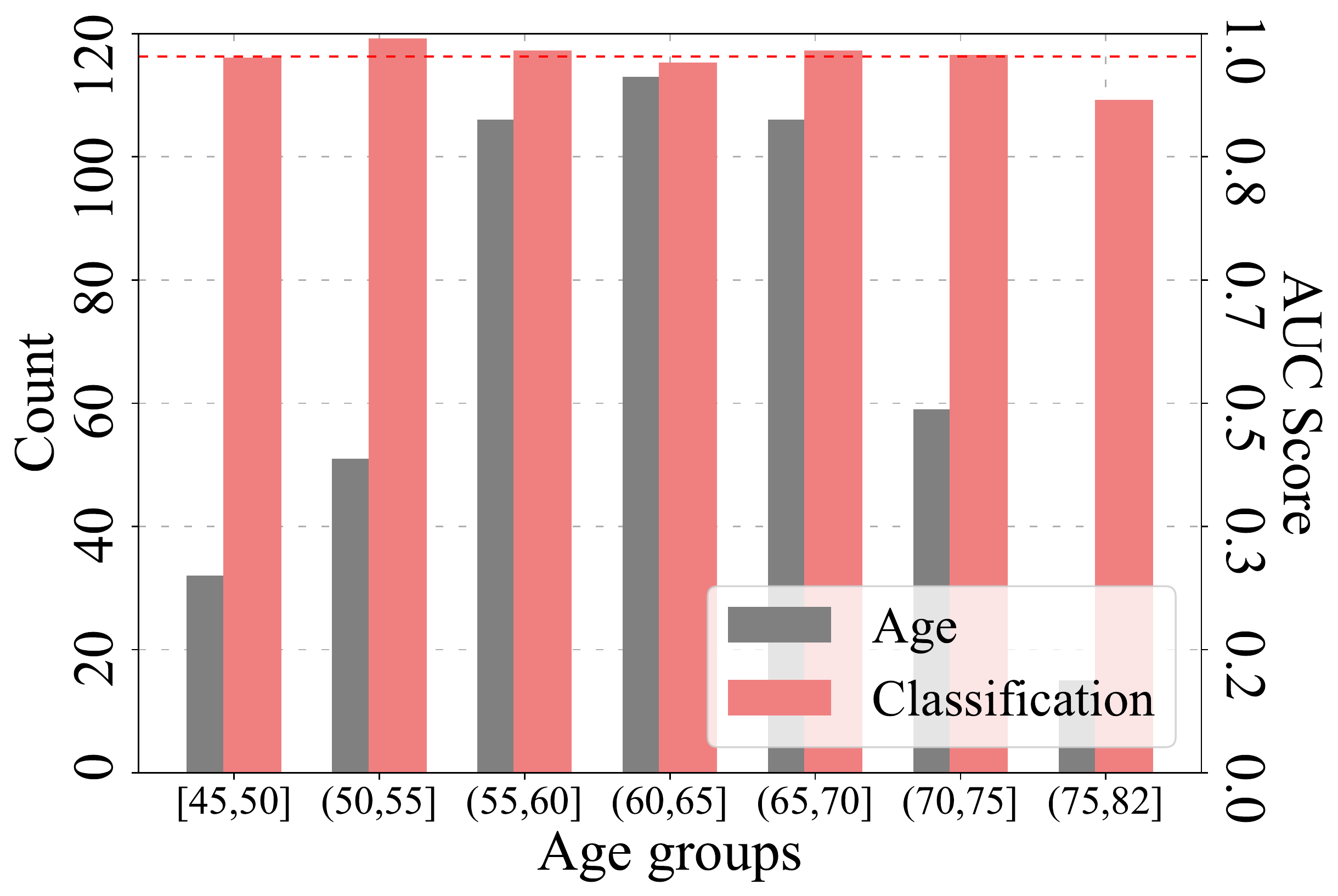}
\endminipage\hfill
\minipage{0.24\textwidth}
  \includegraphics[width=\linewidth]{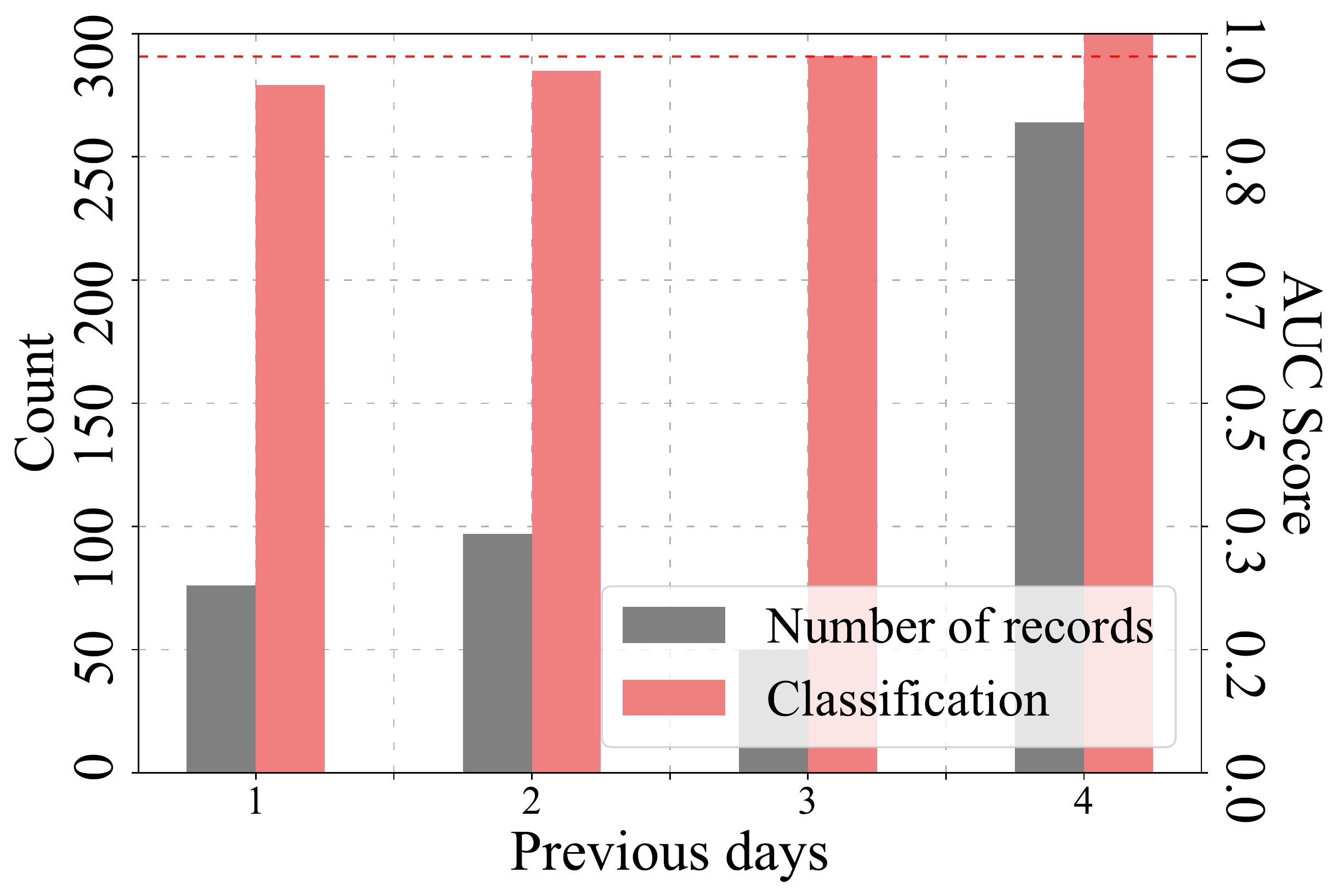}
\endminipage\hfill
\caption{AUC scores on different groups of participants. Left: on different age groups. Right: on different groups with different number of same-labeled historical records. Gray bars: number of participants. Red bars: AUC scores.}
\label{fig:age_prev_auc}
\vspace{-2mm}
\end{figure}

In reality, participant have different demographics and testing preferences. We were curious on how our model performs on different subgroups of people. We divide participants into different groups and examine the performance of our method on each group. First, we evaluate our method on seven age groups ranged from [45,50] to (75,82] with 5 years as a threshold as is shown in Figure~\ref{fig:age_prev_auc}. The results show that our method achieve stable performance (AUC: 0.96$\pm$0.01) across seven age groups even though the number of participants varies in age groups (Count: 68.8$\pm$36.6). This indicate that age poses negligible effects on our method. We also investigate the influence of the number of the same-labeled records in history on the prediction of the incoming record. As can be seen from Figure~\ref{fig:age_prev_auc}, as the same-labeled records increase in the memory, the performance of our method increase steadily and finally reaches AUC=1.0 when all the records are labeled the same as the incoming record. Our method is moderately biased toward samples with more same-labeled records. Nevertheless, the method is still able to achieve reasonable predictions (AUC=0.93) when there is only 1 record in the memory indicating the method's stability.

\newpage
\bibliographystyle{./IEEEtran}
\bibliography{./IEEEexample}
\end{document}